\documentclass[letterpaper, 10 pt, conference]{ieeeconf}  % Comment this line out if you need a4paper

\IEEEoverridecommandlockouts                              % This command is only needed if 
\usepackage{cite}
\usepackage{amsmath,amssymb,amsfonts}
\usepackage{algorithmic}
\usepackage{graphicx}
\usepackage{textcomp}
\usepackage[dvipsnames,x11names]{xcolor}

\usepackage{times}
\usepackage{epsfig}
\usepackage{graphicx}
\usepackage{amsmath}
\usepackage{amssymb}
\usepackage{textcomp}
\usepackage{multirow}
\usepackage{adjustbox}

\usepackage{multirow}
\usepackage{colortbl}
\usepackage{hhline}
\usepackage{subcaption}
\usepackage{verbatim}
\usepackage{tabularx}
\usepackage[group-separator={,},mode=text]{siunitx}

\let\labelindent\relax
\usepackage{enumitem}

\usepackage{gensymb}
\usepackage{flushend}
\usepackage{xcolor}

\usepackage[ruled,vlined]{algorithm2e}

\usepackage[labelformat=simple]{subcaption}
\DeclareCaptionLabelSeparator{periodspace}{.\quad}
\definecolor{green1}{RGB}{14, 81, 7}
\definecolor{green2}{RGB}{63, 125, 49}
\definecolor{green3}{RGB}{123, 175, 112}

\renewcommand{\thetable}{\arabic{table}}
\usepackage{booktabs}

\usepackage{array}

\def\BibTeX{{\rm B\kern-.05em{\sc i\kern-.025em b}\kern-.08em
    T\kern-.1667em\lower.7ex\hbox{E}\kern-.125emX\}}}

\makeatletter
\let\NAT@parse\undefined
\makeatother
\usepackage{hyperref}
\hypersetup{
  colorlinks=true,
  citecolor=SpringGreen4
}
\usepackage{cleveref}

\title{\LARGE \bf
Woodscape Fisheye Semantic Segmentation for Autonomous Driving | CVPR 2021 OmniCV Workshop Challenge
}

\author{Saravanabalagi Ramachandran*, Ganesh Sistu$^{+}$, John McDonald*, and Senthil Yogamani$^{+}$% <-this % stops a space
\thanks{*Saravanabalagi Ramachandran and John McDonald are with Lero - the Irish Software Research Centre and the Department of Computer Science, Maynooth University, Maynooth, Ireland {\tt \{saravanabalagi.ramachandran, john.mcdonald\}@mu.ie}}
\thanks{$^{+}$Ganesh Sistu and Senthil Yogamani are with Valeo Vision Systems, Ireland {\tt \{ganesh.sistu, senthil.yogamani\}@valeo.com}}
}

\begin{document}

\maketitle
\thispagestyle{empty}
\pagestyle{empty}

\begin{abstract}
We present the WoodScape fisheye semantic segmentation challenge for autonomous driving which was held as part of the CVPR 2021 Workshop on Omnidirectional Computer Vision (OmniCV). This challenge is one of the first opportunities for the research community to evaluate the semantic segmentation techniques targeted for fisheye camera perception. Due to strong radial distortion standard models don't generalize well to fisheye images and hence the deformations in the visual appearance of objects and entities needs to be encoded implicitly or as explicit knowledge. This challenge served as a medium to investigate the challenges and new methodologies to handle the complexities with perception on fisheye images. The challenge was hosted on CodaLab and used the recently released WoodScape dataset comprising of 10k samples. 
In this paper, we provide a summary of the competition which attracted the participation of 71 global teams and a total of 395 submissions. The top teams recorded significantly improved mean IoU and accuracy scores over the baseline PSPNet with ResNet-50 backbone. We summarize the methods of winning algorithms and analyze the failure cases. We conclude by providing future directions for the research.
\end{abstract}
\section{Introduction}

In Autonomous Driving, Near Field is a region from 0-15 meters and 360° coverage around the vehicle. Near Field perception is primarily needed for use cases such as automated parking, traffic jam assist, and urban driving  where the predominant sensor suite includes ultrasonics \cite{popperli2019capsule} and surround view fisheye-cameras. Despite the importance of such use cases, the majority of research to date has focused on far-field perception and, as a consequence, there are limited datasets and research on near-field perception tasks. In contrast to far-field, near-field perception is more challenging due to high precision object detection requirements of 10 cm \cite{eising2021near}. For example, an autonomous car needs to be parked in a tight space where high precision detection is required with no room for error.

\begin{figure}[tb]
\centering
\includegraphics[width=\columnwidth]{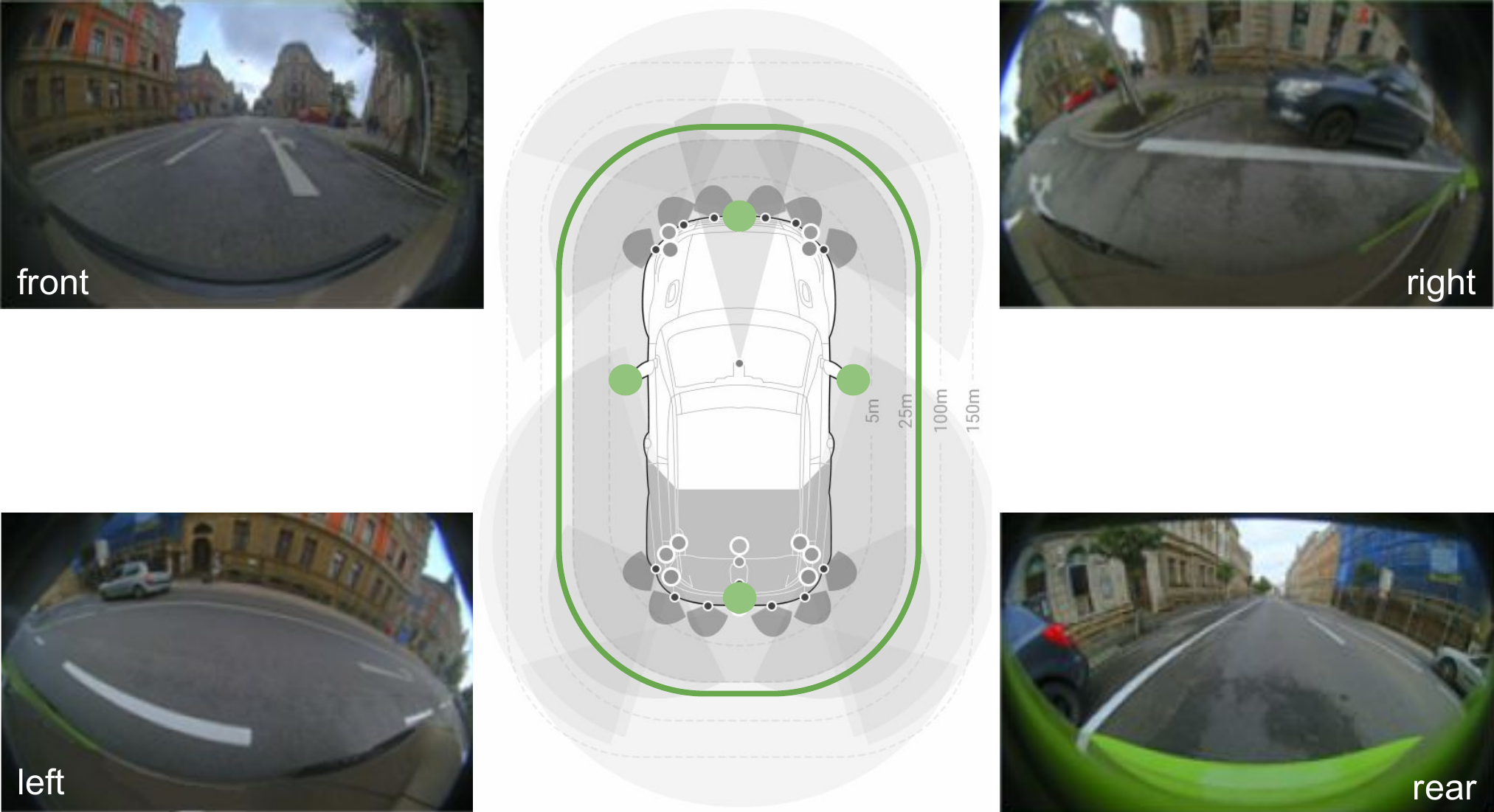}
\caption{Illustration of the four fisheye cameras around the vehicle providing 360$^\circ$ coverage.} 
\label{fig:svs}
\end{figure}

Surround view cameras consisting of four fisheye cameras are sufficient to cover the near-field perception as shown in \autoref{fig:svs}. Standard algorithms can not be extended easily on fisheye cameras due to their large radial distortion. Most algorithms are usually designed to work on rectified pinhole camera images. The naive approach to operating on fisheye images is to first rectify the images and then directly apply these standard algorithms. However, such an approach carries significant drawbacks due to the reduced field-of-view and resampling distortion artefacts in the periphery of the rectified images.

Fisheye cameras are a primary sensor available in most commercial vehicles for automated parking. Even in lower cost vehicles, rear-view fisheye cameras have became a standard feature for dashboard viewing and reverse parking. Fisheye cameras are also used commonly in other domains like video surveillance \cite{kim2016fisheye} and augmented reality \cite{schmalstieg2016augmented}. In spite of its prevalence, there are only a few public datasets for fisheye images publicly available and thus relatively little research is performed.  The Oxford Robotcar dataset \cite{maddern20171} is one such dataset providing fisheye camera images for autonomous driving. It contains over 100 repetitions of a consistent route through Oxford, UK, captured over a period of over a year and used widely for long-term localisation and mapping. OmniScape \cite{sekkat2020omniscape} is a synthetic dataset providing semantic segmentation annotations for cameras mounted on a motorcycle.

\begin{figure*}[tb]
\centering
\includegraphics[width=\textwidth]{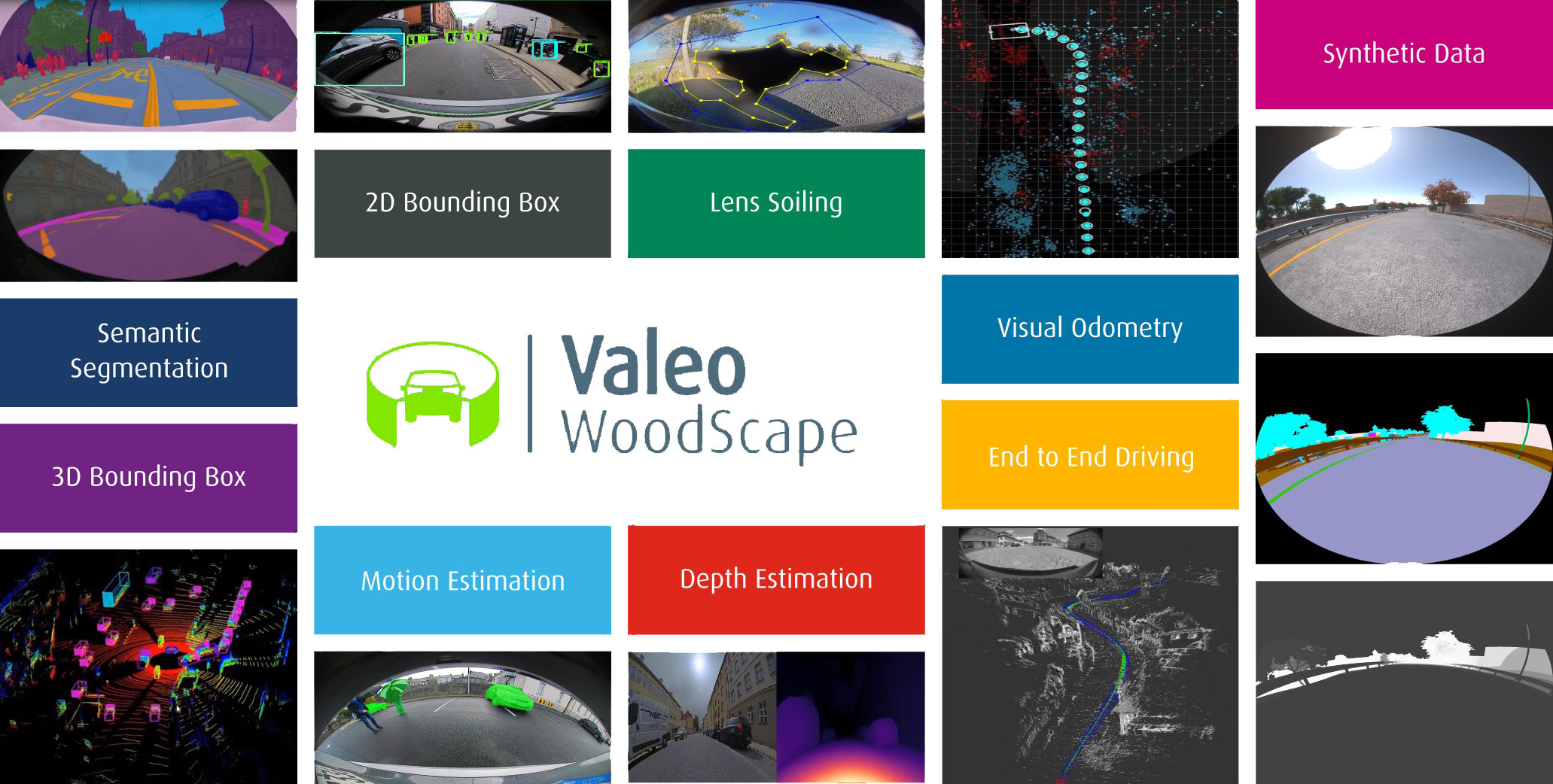}
\caption{Overview of WoodScape dataset tasks.} 
\label{fig:woodscape-tasks}
\end{figure*}

WoodScape \cite{yogamani2019_woodscape} is the world’s first public surround view fisheye automotive dataset released to accelerate research in multi-task multi-camera computer vision for automated driving. The dataset sensor setup comprises of four surround-view fisheye cameras covering $360^\degree$ around the vehicle. The dataset consists of annotations for nine tasks, including segmentation, depth estimation, bounding boxes, pixel level motion masks and a novel lens soiling detection task (illustrated in \autoref{fig:woodscape-tasks}). Semantic instances for 40+ classes provided for over \num{10000} images. WoodScape dataset encourages researchers to design algorithms that operate directly on fisheye images, modelling the inherent distortion, rather than using naive rectification. The dataset has enabled such research for depth estimation \cite{kumar2020fisheyedistancenet}, object detection \cite{rashed2021generalized},  soiling detection \cite{uricar2021let}, trailer detection \cite{dahal2019deeptrailerassist} and multi-task learning \cite{kumar2021omnidet}. In this paper, we discuss our challenge focused on the semantic segmentation task in WoodScape.

\section{Challenge}

\begin{figure*}[tb]
\centering
\includegraphics[width=\textwidth]{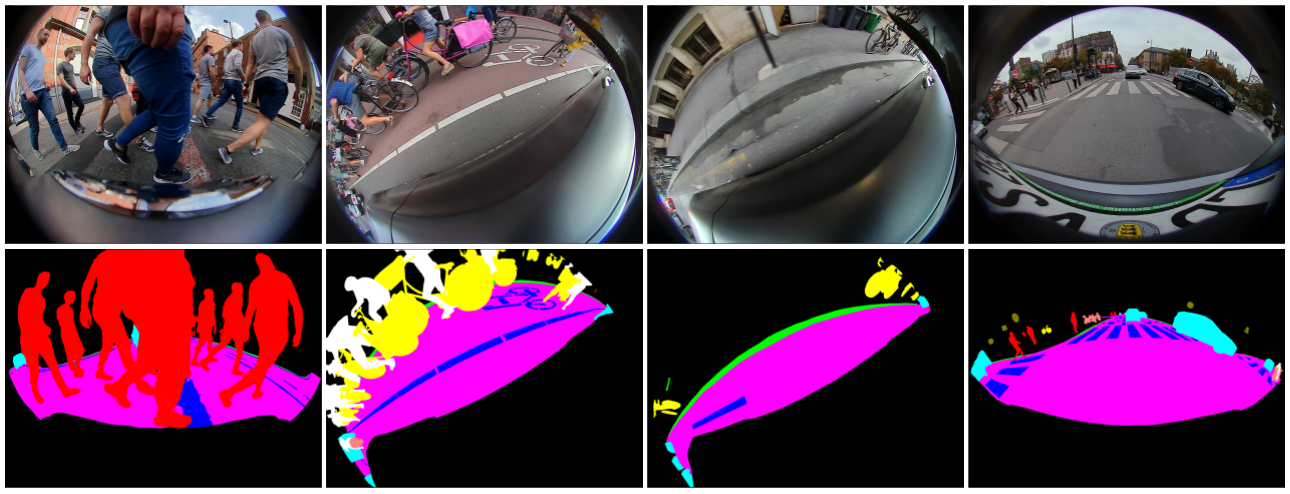}
\caption{Illustration of semantic segmentation annotations.} 
\label{fig:segmentation-task}
\end{figure*}

The objective of the challenge is  to benchmark fisheye semantic segmentation techniques encouraging modelling of radial distortion either implicitly or explicitly using calibration.
Sample images and corresponding annotations of our segmentation dataset are shown in \autoref{fig:segmentation-task}. Compared to other driving datasets, one can observe heavy distortions with near field objects due to fisheye lens property.

Dataset statistics are shown in \autoref{tab:train_test_split}. The ground truth labels consist of 10 classes including void (class 0) and the 9 other classes explained in \autoref{tab:label_mapping}. Although the original dataset consists of over 40 annotated classes with individual instances, they are aggregated or eliminated into 10 important classes which are necessary for autonomous driving applications.
\begin{table}
\centering
\begin{tabular}{ccc} \toprule
Split & Images & Percent \\\midrule
Training Set & 8234 & 82.34\% \\
Test Set & 1766 & 17.66\% \\\midrule
Total & 10000 & 100.00\% \\\bottomrule
\end{tabular}
\quad

\caption{Table showing the split of images between the train and test sets.}
\label{tab:train_test_split}
\end{table}

\begin{table}
\centering
\begin{tabular}{ccc} \toprule
Label & Description \\\midrule
0 & Void \\
1 & Road \\
2 & Lanemarks \\
3 & Curb \\
4 & Pedestrians \\\bottomrule
\end{tabular}
\quad
\begin{tabular}{ccc} \toprule
Label & Description \\\midrule
5 & Rider \\
6 & Vehicles \\
7 & Bicycle \\
8 & Motorcycle \\
9 & Traffic Sign \\\bottomrule
\end{tabular}
\caption{Table showing the description of labels 0-9.}
\label{tab:label_mapping}
\end{table}

\subsection{Metrics}

Mean Intersection-over-Union (mIoU) is a standard evaluation metric for semantic image segmentation, which first computes the Intersection-over-Union (IoU) for each semantic class and then computes the average over classes. IoU is defined as follows: 

\begin{equation}
\text{IoU} = \frac{\text{TP}}{\text{TP} + \text{FP} + \text{FN}}
\end{equation}
where TP is True Positive, FP is False Positive and FN is False Negative.
% \begin{conditions}
%     \text{TP} & True Positive \\
%     \text{FP} & False Positive \\
%     \text{FN} & False Negative \\
% \end{conditions}

The competition entries in CodaLab were evaluated and ordered based the mean Intersection over Union (mIoU) score applied to all classes except class 0 - void class, to give more priority to non-void classes. We specify this metric as \textit{Score} in the leaderboard. The leaderboard also displayed mIoU and mean accuaracy (mAcc) over all the classes, however, these scores were provided for information purposes only and were not used in the overall ranking. 

\subsection{Conditions}

In this competition participants were allowed to use public datasets for domain adaptation or pre-training. There were no limits on training time or network capacity. Individuals and teams (of any size) were allowed to enter the competition, with limits of 10 submissions per day and 100 submissions in total per person/team. Valeo employees or employees of third party companies, universities or institutions that contributed to the creation or have access to the full WoodScape dataset were not allowed to take part in this challenge.

\section{Outcome}

\begin{figure*}[tb]
\centering
\includegraphics[width=\textwidth]{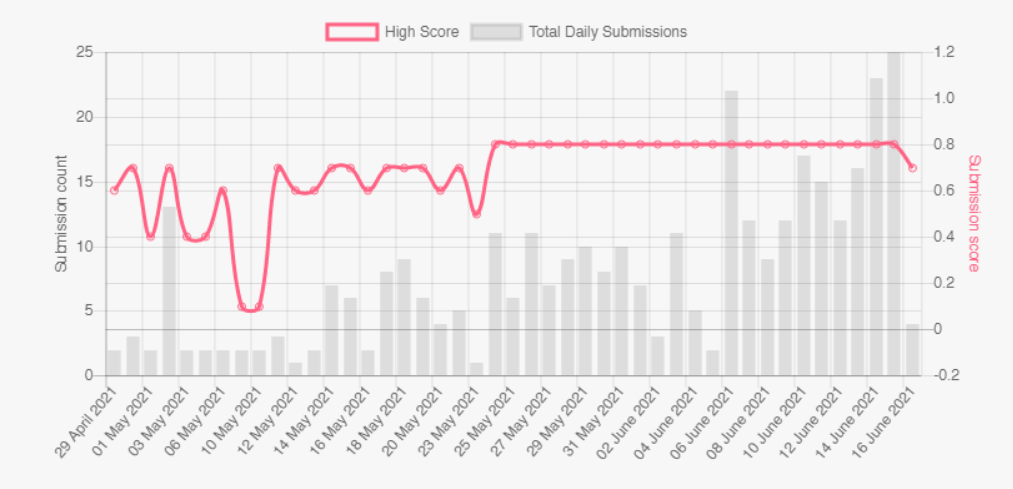}
\caption{Graph showing number of daily submissions and submission scores during all the 55 days of the competition.} 
\label{fig:analytics}
\end{figure*}

The competition was active for 55 days from April 23, 2021 00:00 through to April June 16, 2021 23:59. The competition attracted a total of 71 participants with 395 submissions. A graph showing submissions and submission scores on a day to day basis is shown in \autoref{fig:analytics}. It can be seen from the graph that during the first half we observed between 0 and 7 daily submissions, while during the second half it ranged from 7 to 20 with some exceptions in both. Daily submissions made in the second half consistently had at least one submission with score greater than $0.75$. Overall, the competition attracted approximately 7 submissions per day.

\subsection{Methods}

\subsubsection{Baseline}
In order to provide a baseline performance score a PSPNet\cite{zhao2017pyramid_pspnet} network was provided with a ResNet-50\cite{he2015deep_resnet} backbone finetuned on WoodScape Dataset \cite{yogamani2019_woodscape} (trained for 40,000 iterations). Cross validation  and data augmentation techniques were not employed. The baseline network achieved a score of 0.56 (mIoU 0.50, accuracy 0.67) excluding void class. 
It is notable that PSPNet had reported state of the art scores on PASCAL VOC2012 and Cityscapes dataset at the time of its release in 2017 and served as a good baseline for semantic segmentation tasks at the time of writing. However, even after fine-tuning we observed that there is a significant room for improvements for it to be able to perform better on the WoodScape dataset. We infer that this is due to the high radial distortions present in the fisheye images, while the network was pretrained on ImageNet \cite{deng2009_imagenet} which does not contain images with such distortion. Thus, we see a big potential for the competition participants to demonstrate substantially improved performance over the baseline.\\

\begin{figure*}[tb]
\centering
\includegraphics[width=\textwidth]{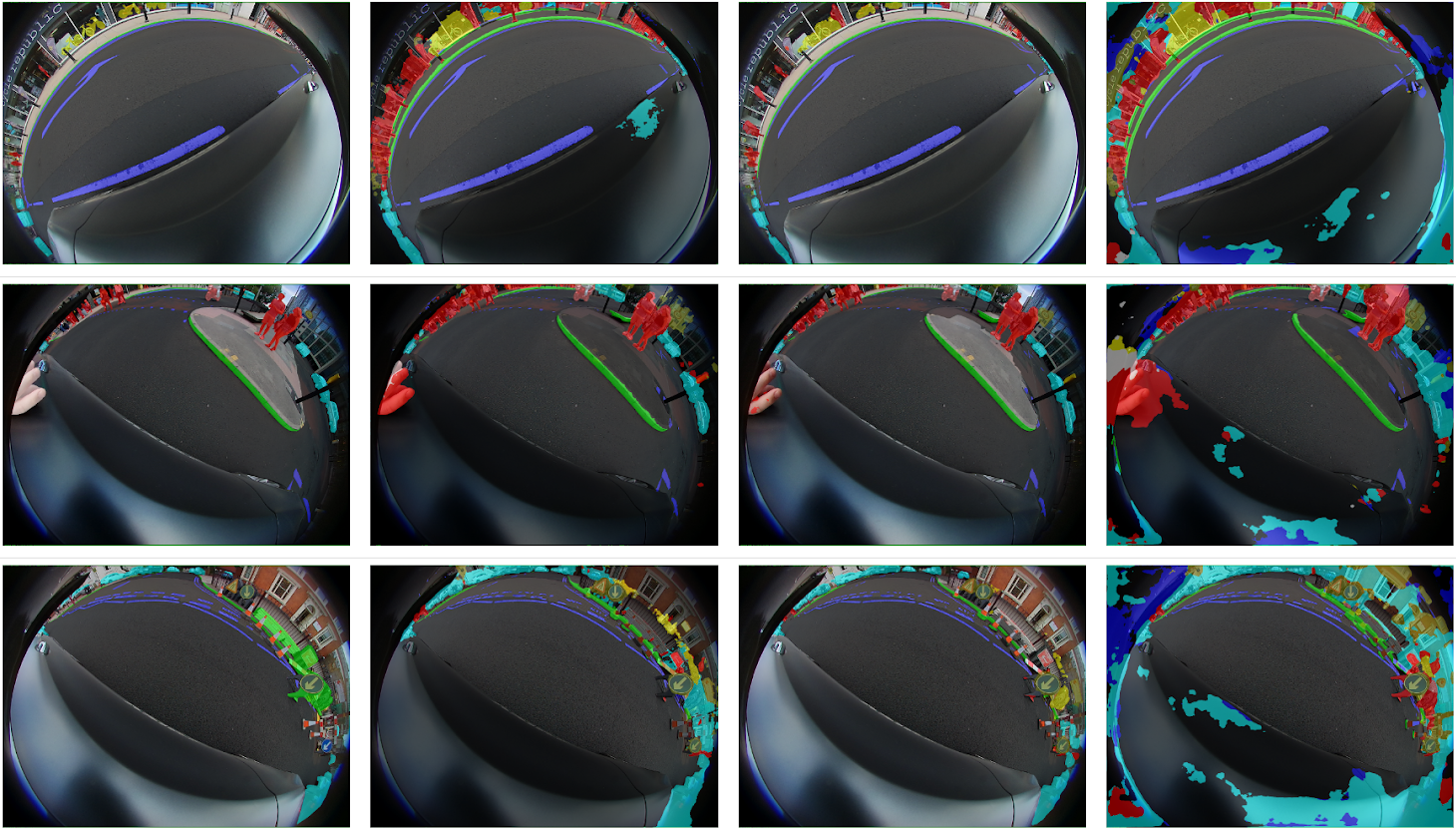}
\caption{Semantic segmentation masks predicted by top 3 teams compared against the ground truth masks. Label 0 and 1 are excluded to make images easy to perceive. Left to Right: Ground Truth and prediction results of winner (earhian), second (soeaver) and third place (raman\_focus) teams respectively.} 
\label{fig:pred-comparison}
\end{figure*}

\subsubsection{Winning Team}
Team Meituan finished in first place with a score of 0.84 (mIoU 0.86, accuracy 0.89) with their full Swin-transformer Encoder-Decoder approach. 
Jian Qiao, 
Haichao Shi, 
Xinchu Shi, 
Bocong Liu, and 
Xiaoyu Zhang, 
affiliated to 
Autonomous Delivery of Meituan, and 
Institute of Information Engineering, Chinese Academy of Sciences belonged to this team.
They adopted Swin-Transformer\cite{liu2021swin} to create two models. The first model is based on pure Swin-Transformer, where the PixelShuffle \cite{shi2016realtime_pixelshuffle} operation is used in the up-sampling layer of decoder. The second model was constructed using the encoder of Swin-Transformer, and deeplabv3 as the decoder. In the training stage, the two models used Mixup \cite{zhang2018mixup}, EMA \cite{he2020momentum_moco}, cross-validation, multi-scale training and etc. In the testing stage, a voting strategy is used to create an ensemble model that combined the two models.\\

\begin{table*}
\centering
\begin{tabular}{clcccc} \toprule
Rank & Username & Team Name         & Score $\uparrow$ & mIoU $\uparrow$ & Accuracy $\uparrow$ \\\midrule
\color{green1}\textbf{1} & \color{green1}\textbf{earhian}  & \color{green1}\textbf{Meituan}              & \color{green1}\textbf{0.84} \ \;(1) & \color{green1}\textbf{0.86} \ \;(1) & \color{green1}\textbf{0.91} \ \;(1) \\
\color{green2}\textbf{2} & \color{green2}\textbf{HaichaoShi} & \color{green1}\textbf{Meituan}            & \color{green2}\textbf{0.83} \ \;(2) & \color{green2}\textbf{0.85} \ \;(2) & \color{green3}\textbf{0.90} \ \;(3) \\
\color{green3}\textbf{3} & \color{green3}\textbf{soeaver}  & \color{green2}\textbf{BUPT-PRIV}             & \color{green3}\textbf{0.83} \ \;(3) & 0.75 \ \;(7) & 0.89 \ \;(4) \\
4 & raman\_focus & \color{green3}\textbf{VinAIResearch}    & 0.81 \ \;(4) & \color{green3}\textbf{0.83} \ \;(3) & \color{green2}\textbf{0.90} \ \;(2) \\
5 & DeepBlueAI & DeepBlueAI         & 0.80 \ \;(5) & 0.82 \ \;(4) & 0.89 \ \;(5) \\
6 & fanzonghao &                    & 0.79 \ \;(6) & 0.81 \ \;(5) & 0.88 \ \;(6) \\
7 & TheRealG & The Human Overlords  & 0.77 \ \;(7) & 0.79 \ \;(6) & 0.87 \ \;(7) \\
8 & yangdonghan50 &                 & 0.76 \ \;(8) & 0.68 (10) & 0.86 \ \;(8) \\
9 & tfzhou.ai &                     & 0.72 \ \;(9) & 0.65 (11) & 0.78 (12) \\
10 & install83  &                   & 0.71 (10) & 0.74 \ \;(8) & 0.81 (10) \\
11 & jesse1029 &  & 0.70 (11) & 0.73 \ \;(9) & 0.83 \ \;(9) \\
12 & sgzqc & CAIC\_AI & 0.65 (12) & 0.59 (12) & 0.73 (13) \\
13 & lxixi0815 &  & 0.65 (13) & 0.58 (13) & 0.73 (14) \\
14 & alright1117 &  & 0.64 (14) & 0.57 (14) & 0.72 (15) \\
15 & hpysr\_s &  & 0.60 (15) & 0.54 (15) & 0.79 (11) \\
\color{Brown}\textbf{16} & \color{Brown}\textbf{Baseline} &   & \color{Brown} \textbf{0.56} (16) & \color{Brown} \textbf{0.50} (16) & \color{Brown} \textbf{0.67} (17) \\
17 & stephancheng &  & 0.54 (17) & 0.48 (17) & 0.68 (16) \\
18 & Y\_C &  & 0.48 (18) & 0.44 (18) & 0.59 (18) \\
19 & sabarinathan & Image LAB & 0.45 (19) & 0.40 (19) & 0.54 (21) \\
20 & RobertWood &  & 0.44 (20) & 0.39 (20) & 0.58 (19) \\\bottomrule
\end{tabular}
\caption{Leaderboard showing top-10 user submissions based on the \textit{Score} metric along with baseline score. Top 3 scores in each column are highlighted in shades of green with dark green being first rank. The column-wise rank for the corresponding metric is shown in brackets for \textit{Score}, \textit{mIoU}, and \textit{Accuracy}. Baseline score is shown in brown. }
\label{tab:leaderboard}
\end{table*}

\subsubsection{Second Place}
Team BUPT-PRIV finished in second place with a score of 0.83 (mIoU 0.75, accuracy 0.89) using dense transformers.
Lu  Yang,
Qing Song,
Donghan Yang,
Tianfei Zhou,
Wenguan Wang, and
Liulei Li
affiliated to 
BUPT, ETH Zurich, and BIT belonged to this team. 
They proposed a transformer network for dense pixel prediction, which adopts SWIN\cite{liu2021swin} as the encoder. They also proposed a decoder that can generate high-resolution features by making use of the high-resolution module of HRNet \cite{wang2020deep_hrnet}. The team used a WeightedCrossEntropy loss to cater for the imbalance in object categories. Their model reported a 83\% mIoU score on the test set.\\

\subsubsection{Third Place}
Team VinAIResearch finished in third place with a score of 0.81 (mIoU 0.83, accuracy 0.90) using a multi-head network for multi-view fisheye image segmentation. 
Ahmed Rida Sekkat,
Anh Vo Tran Hai,
Tuan Ho,
Tin Duong, and
Bac Nguyen
affiliated to
VinAiResearch belonged to this team. 
They utilized DeepLabV3+ \cite{wang2020deep_hrnet} architecture with a modified multi-headed attention model and a weighted loss function to obtain a score of 0.80. An ensemble architecture of the two best models further increased their score to 0.81.\\

% \subsubsection{Other interesting approaches}
% \sr{should we include this section given we have not had the list of all approaches}

% Spherical model where they project the data onto a sphere using the detailed calibration data and perform semantic segmentation.

\subsection{Results and Discussion}

Team Meituan, with a lead score of 0.84 (mIoU 0.87, accuracy 0.89), was announced as the winner on 18th June 2021. The leaderboard for the competition showing the top-20 user submissions along with the team names is presented in \autoref{tab:leaderboard}.

% \sr{needs review} 
\autoref{fig:pred-comparison} shows predictions for 3 randomly chosen images from the top three teams.
% It was observed that in some cases the participants including the over winner managed to increase the score by a small margin by opting to not predict \textit{Void} (Label 0) at all, instead replacing it with closest prediction from the rest of the labels (Label 1-9). This can be explained by the fact that, (i) the mean IoU scores are based only on classes 1 to 9 leaving out \textit{Void}, and (ii) that \textit{Road} (Label 1) occupies a significantly larger portion than the rest of the classes in most images, while still being equally weighed in the score calculation.
We chose to not include \textit{Void} to give more importance to other classes, however this has inadvertently enabled the models to take a smaller score penalty on misclassifying \textit{Void} as \textit{Road}. We have excluded both \textit{Void} and \textit{Road} in the comparison for better visual understanding of predictions of other classes.

\autoref{tab:classwise_scores_toppers} shows accuracy and IoU for each class for the top 3 teams. The top 3 teams managed to get IoU scores over 0.9 for \textit{Road}, \textit{Pedestrians} (Label 4), and \textit{Vehicles} (Label 6). This can be attributed to the existence of better performing and refined object detectors for these classes. There were many mispredictions on \textit{Riders} (Label 5) from all the 3 teams with an IoU score of less than 0.7 as can be seen in \autoref{tab:classwise_scores_toppers}. In particular, we notice that many \textit{Rider} pixels are largely misclassified as \textit{Pedestrian}, as riders and pedestrian are very similar in visual appearance and often very difficult to disambiguate from single images (e.g. due to occlusion, etc.).

In future iterations of the challenge, to mitigate such unintended effects on the score, we plan to use weighted averaging of the IoU of classes while including more classes (e.g. buildings, traffic lights, etc.) limiting the \textit{Void} to sky and other miscellaneous objects.

\begin{table}
\centering
\begin{tabular}{llcccccc} \toprule
L & Description & \multicolumn{2}{c}{earhian} & \multicolumn{2}{c}{soeaver} & \multicolumn{2}{c}{raman\_focus} \\
& & Acc $\uparrow$ & IoU $\uparrow$ & Acc $\uparrow$ & IoU $\uparrow$ & Acc $\uparrow$ & IoU $\uparrow$ \\\midrule
1 & Road & \color{green1} \textbf{0.99} & \color{green1} \textbf{0.98} & \color{green1} \textbf{0.99} & \color{green1} \textbf{0.98} & 0.98 & 0.97 \\
2 & Lanemarks & \color{green1} \textbf{0.85} & \color{green1} \textbf{0.76} & 0.83 & \color{green1} \textbf{0.76} & \color{green1} \textbf{0.85} & 0.73 \\
3 & Curb & 0.91 & \color{green1} \textbf{0.82} & 0.87 & 0.81 & \color{green1} \textbf{0.93} & 0.77 \\
4 & Pedestrians & \color{green1} \textbf{0.97} & \color{green1} \textbf{0.95} & 0.95 & 0.93 & \color{green1} \textbf{0.97} & 0.92 \\
5 & Rider & \color{green1} \textbf{0.68} & \color{green1} \textbf{0.57} & 0.58 & 0.51 & 0.61 & 0.55 \\
6 & Vehicles & 0.98 & \color{green1} \textbf{0.97} & 0.98 & 0.96 & \color{green1} \textbf{0.99} & 0.96 \\
7 & Bicycle & 0.90 & 0.84 & \color{green1} \textbf{0.93} & \color{green1} \textbf{0.86} & \color{green1} \textbf{0.93} & 0.82 \\
8 & Motorcycle & 0.87 & \color{green1} \textbf{0.82} & \color{green1} \textbf{0.88} & 0.79 & \color{green1} \textbf{0.88} & 0.73 \\
9 & Traffic Sign & \color{green1} \textbf{0.90} & \color{green1} \textbf{0.87} & \color{green1} \textbf{0.90} & 0.86 & 0.86 & 0.80 \\\bottomrule
\end{tabular}
\caption{Table showing class-wise accuracy and IoU scores for the top three teams. Best of the 3 Accuracy and IoU is highlighted in dark green.}
\label{tab:classwise_scores_toppers}
\end{table}

% \sy{wud be good to analyze failure cases and add some insights for future work}
\section{Conclusion}
In this paper, we discussed the results of the first fisheye semantic segmentation challenge hosted at our CVPR OmniCV workshop 2021. Fisheye cameras have a spatially variant distortion which makes it challenging for CNNs as its translation invariance breaks. The top methods made use of transformers which have a spatial adaptation mechanism implicitly at patch level and hence avoided the explicit modelling of the radial distortion. However, incorporating the known camera geometry would have acted as an inductive bias for efficient learning and better generalization. In future work, we plan to organize similar workshop challenges on panoptic segmentation.

\section*{Acknowledgments}

Woodscape OmniCV 2021 Challenge
% presented in this paper
was supported
in part
by \href{https://www.sfi.ie}{Science Foundation Ireland} grant 13/RC/2094 to \href{https://www.lero.ie}{Lero - the Irish Software Research Centre} and grant 16/RI/3399.

\bibliographystyle{ieeetran}
\bibliography{IEEEabrv, references}

\end{document}